\pdfoutput=1

\documentclass[11pt]{article}

\usepackage{acl}

\usepackage{times}
\usepackage{latexsym}

\usepackage[T1]{fontenc}

\usepackage[utf8]{inputenc}

\usepackage{microtype}

\usepackage{inconsolata}

\usepackage{graphicx}
\usepackage{amsmath}
\usepackage{ifthen}
\usepackage{algorithm}
\usepackage{multirow}    
\usepackage{array}       
\usepackage{xcolor}
\definecolor{seagreen}{rgb}{0.180, 0.545, 0.341}
\usepackage{booktabs}
\usepackage{color}
\usepackage{longtable}
\usepackage{array}       
\usepackage{longtable}   
\usepackage{xltabular}   
\usepackage{booktabs}    
\usepackage{makecell}    
\usepackage{subcaption}
\usepackage{bm}
\usepackage{arydshln} 
\usepackage{threeparttable}
\usepackage{microtype}
\usepackage{tabularx}
\title{Watch Closely: Mitigating Object Hallucinations in Large Vision-Language Models with Disentangled Decoding}

\author{
  Ruiqi Ma$^1$, 
  Yu Yan$^1$, 
  Chunhong Zhang$^1$, 
  Minghao Yin$^2$, \\
  {\bf XinChao Liu$^1$, 
  Zhihong Jin$^1$, 
  Zheng Hu$^1$} \\
  $^1$Beijing University of Posts and Telecommunications \\
  $^2$The University of Hong Kong
}

\begin{document}
\maketitle
\begin{abstract}
Large Vision-Language Models (LVLMs) bridge the gap between visual and linguistic modalities, demonstrating strong potential across a variety of domains. However, despite significant progress, LVLMs still suffer from severe hallucination issues in object recognition tasks. These models often fail to accurately identify certain objects, leading to text generation that appears fluent but does not correspond to the visual content, which can have serious consequences in real-world applications. Recently, several methods have been proposed to alleviate LVLM hallucinations, but most focus solely on reducing hallucinations in the language modality. To mitigate hallucinations in both the language and visual modalities, we introduce Hallucination Disentangled Decoding (HDD) method that requires no training. HDD enhances the original image by segmenting it and selecting images that augment the original, while also utilizing a blank image to eliminate language prior hallucinations in both the original and segmented images. This design not only reduces the model's dependence on language priors but also enhances its visual performance. (Code: \url{https://github.com/rickeyhhh/Hallucination-Disentangled-Decoding}) 
\end{abstract}

\section{Introduction}
Large Vision-Language Models (LVLMs) \cite{Yin2023ASO} typically consist of a large language model and a visual encoder or multimodal vision model. The visual module effectively captures feature information from images and maps this visual information to text in a shared space \cite{Dosovitskiy2020AnII}. Leveraging their remarkable image processing capabilities, LVLMs have already demonstrated significant value across various domains. However, challenges remain, with one of the most prominent being the issue of multimodal hallucination. The multimodal hallucination problem in LVLMs is primarily manifested in the inability of the text output to accurately correspond to the input image information, including the generation of non-existent objects and errors in object recognition. This issue undermines the model's reliability to some extent.

Some studies have already investigated "how hallucinations arise," revealing the LVLM's potential over-reliance on the language priors of the backbone language model \cite{yue-etal-2024-less} and biases in the training datasets \cite{Liu2024ASO}. Recently, many approaches have been proposed to alleviate such multimodal hallucinations, including manually constructing negative samples for contrastive decoding \cite{Leng2023MitigatingOH}, identifying summary tokens generated during LVLM output and applying penalty mechanisms \cite{Huang2023OPERAAH}, as well as training methods that encourage the generation of EOS tokens to prevent the model from producing overly detailed or hallucinated content \cite{yue-etal-2024-less}. While these methods have alleviated LVLM hallucinations to some extent, multimodal hallucination remains an entangled phenomenon, originating from both of the LVLM's main components: the visual and language modules, each of which generates its own type of hallucination. The visual module has limitations in its perceptual capabilities and may fail to accurately recognize smaller objects in highly informative images \cite{Kan2024CATCHCA}. The language module, due to its autoregressive nature, tends to emphasize linguistic coherence \cite{Liu2024PayingMA} when generating the next token, rather than ensuring correspondence between the text and the image. The aforementioned methods have not disentangled the sources of LVLM hallucinations, but rather address hallucinations in a coarse-grained manner, which can lead to a performance trade-off when alleviating hallucinations in one module at the cost of another. To address this issue, we proposes a approach by disentangling the analysis of the sources of hallucinations in the LVLM's visual and language modules. 

In this work, we designed experiments to analyze the sensitivity of LVLMs to local image details, leading to the conclusion that the sensitivity of the LVLM's visual encoder to entities in the image is correlated with the size of the entities \cite{Zhang2024ExploringPL}. During the course of these experiments, we also identified and analyzed the LVLM's dependency on language priors, which aligns with previous research findings \cite{Wang2024MitigatingHI}. Through this research, we decompose the entangled concept of LVLM hallucinations into two clearer components: visual hallucinations \cite{Guan2023HallusionbenchAA} and language hallucinations \cite{Liu2024ASO}. To address these two sources of hallucinations, we propose a disentangled decoding approach to reduce LVLM hallucinations. We segment the original image and enhance the original image by calculating the Jensen-Shannon divergence \cite{fuglede2004jensen} between the segmented image and the original image, as well as contrastive decoding \cite{Wang2024MitigatingHI} using blank images to eliminate the LVLM's reliance on language priors. Our approach is highly effective, outperforming recent methods and baselines across multiple hallucination benchmark tests.

Overall, our contributions are as follows:
\begin{enumerate}
    \item We conducted an in-depth study of the LVLM's sensitivity to local image details and analyzed its excessive reliance on language priors.
    \item We propose the Hallucination Disentangled Decoding method, which overcomes the limitations of previous approaches that were unable to disentangle and eliminate hallucinations from both visual and language modules.
    \item Through various benchmark tests, HDD demonstrates superior perceptual capabilities and generates fewer hallucinations without any additional training or fine-tuning.
\end{enumerate}

\section{Related Work}
\subsection{Hallucinations in LVLMs}
Multimodal hallucinations include object hallucinations, attribute hallucinations, relation hallucinations \cite{bai2024hallucination}, etc. Here we mainly focus on object hallucinations. There are two main causes of LVLM's object hallucinations, (1) Limited perceptual capabilities of the visual module \cite{Liu2024ASO}. For entities occupying a small area in the image, the visual module may struggle to perceive them in detail, leading to either omission or incorrect descriptions. Of course, there are many factors that influence the limitations of the visual module, such as training data contamination \cite{Yu2023HalluciDoctorMH}, overly detailed annotations \cite{yue-etal-2024-less}, and data biases \cite{Liu2024ASO}. (2) Excessive reliance of the language module on the language priors of the backbone LLM \cite{chen2025perturbollavareducingmultimodalhallucinations}. The proportion of image information decreases during the sequence generation process \cite{yue-etal-2024-less}, which is a characteristic of autoregressive models. Hallucinations generated by the visual encoder can induce the language model to generate more hallucinations, causing these two types of hallucinations to often be entangled.

\subsection{Hallucination Mitigation Methods}
\begin{figure}[t]
  \includegraphics[width=\columnwidth]{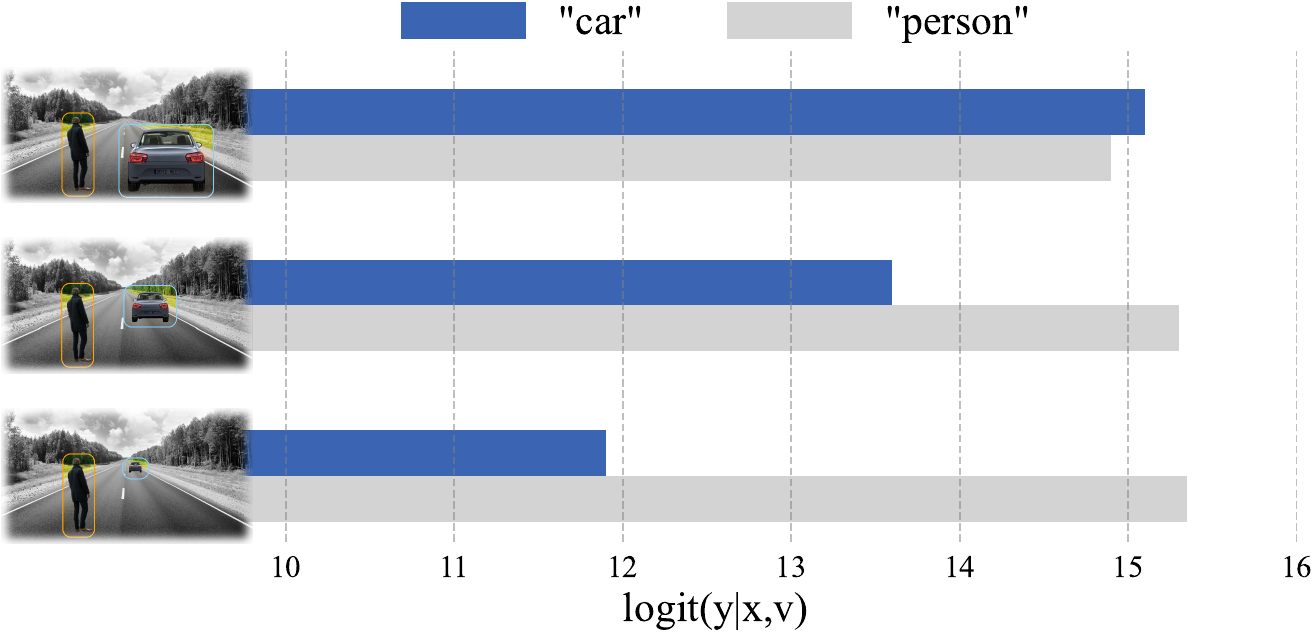}
  \caption{The figure above illustrates the impact of different object sizes in the image on the model's output logits distribution. From top to bottom, the area of the "\textit{car}" gradually decreases, while the area of the "\textit{person}" remains unchanged, serving as a control.}
  \label{fig:figure2}
\end{figure}
Recently, numerous efforts have focused on mitigating object hallucinations in LVLMs through the design of different decoding strategies. OPERA \cite{Huang2023OPERAAH} introduces a decoding approach that adjusts attention weights with a penalty term and incorporates a fallback mechanism to suppress the influence of aggregate tokens during generation. VCD \cite{Leng2023MitigatingOH} introduces visual noise to construct negative samples and employs contrastive decoding to reduce hallucinations. HALC \cite{chen2024halcobjecthallucinationreduction} leverages an auxiliary vision model to identify hallucinated tokens and performs multi-scale resampling over the input image to refine the output distribution. SID \cite{huo2025selfintrospectivedecodingalleviatinghallucinations} mitigates hallucinations via token-level perturbation-induced decoding, while RITUAL \cite{Woo2024RITUALRI} employs randomized image transformations to suppress hallucinated content. In this work, we propose the first disentangled framework for hallucination mitigation in LVLMs. Our approach leverages semantic segmentation model to construct additional structured visual input and employs blank images to eliminate the model’s reliance on language priors, effectively mitigating the hallucinations between visual and language modules. The proposed method is intuitive, highly interpretable, and achieves state-of-the-art performance across various evaluation metrics.
\section{Preliminary}
\subsection{Decoding in LVLMs}
We model the LVLM as a function $\mathcal{M_\theta}$, where the input consists of an image input \(v\) and a text query input \(x\). Mathematically, at the t-th decoding step, the auto-regressive process can be represented as:
\setlength{\abovedisplayskip}{3pt}
\setlength{\belowdisplayskip}{3pt}
\begin{equation}
\small
\label{eq:1}
    y_t=\mathcal{M_\theta}(v,x,y_{<t})
\end{equation}
\(y_{<t}\) represents the results decoded in the previous \(t-1\) steps, where the next token is predicted based on the previously generated tokens. We can expand Equation \ref{eq:1} as follows:
\begin{equation}
\label{eq:3}
    p(y_t|v,x,y_{<t})=\mathrm{SoftMax}[\mathrm{logit}(y_t|v,x,y_{<t})] \small
\end{equation}
\begin{equation}
\label{eq:4}
    y_t \sim p(y_t|v,x,y_{<t}) \small
\end{equation}
For simplicity of representation, the processing of the visual encoder and the text encoder is not shown here. The model generates the logits distribution for the next token based on the image information \(v\), the query \(x\), and the previously generated tokens \(y_{<t}\). This logit distribution is then normalized into a probability distribution via \(\mathrm{SoftMax}\), and a token is selected using specific sampling strategy. Hallucinations can arise at two stages, (1) the local detail bias in the output \(v\) after the image input passes through the visual encoder, and (2) the language prior bias introduced by the iteratively generated preceding tokens \(y_{<t}\). Our method addresses these two challenges separately.

\subsection{Semantic Segmentation Tool}
We construct additional visual information by leveraging semantic segmentation tools to automatically generate entity-level masks and partition the input image accordingly. In our main experiments, we adopt the Segment Anything Model \cite{Kirillov2023SegmentA} as the primary segmentation backbone. For ablation studies, we further compare the effectiveness of our framework when integrated with alternative segmentation models, including Mask2Former \cite{cheng2022maskedattentionmasktransformeruniversal} and Mask R-CNN \cite{he2018maskrcnn}. The partitioning process can be expressed as:
\begin{equation}
\label{eq:5}
    \small
    \mathcal{S}eg(v)=\sum_{i=1}^n(Mask_i\odot v)
\end{equation}
where \(v\) represents the image input to the segmentation model. After processing , the image \(v\) produces \(n\) masks, which are multiplied pixel-wise with the original image. Each result represents a distinct entity, and by combining these entities, a new additional visual image is obtained.
\subsection{Local Detail-Induced Hallucinations}
\begin{figure}[t]
  \raggedleft
  \includegraphics[width=\columnwidth]{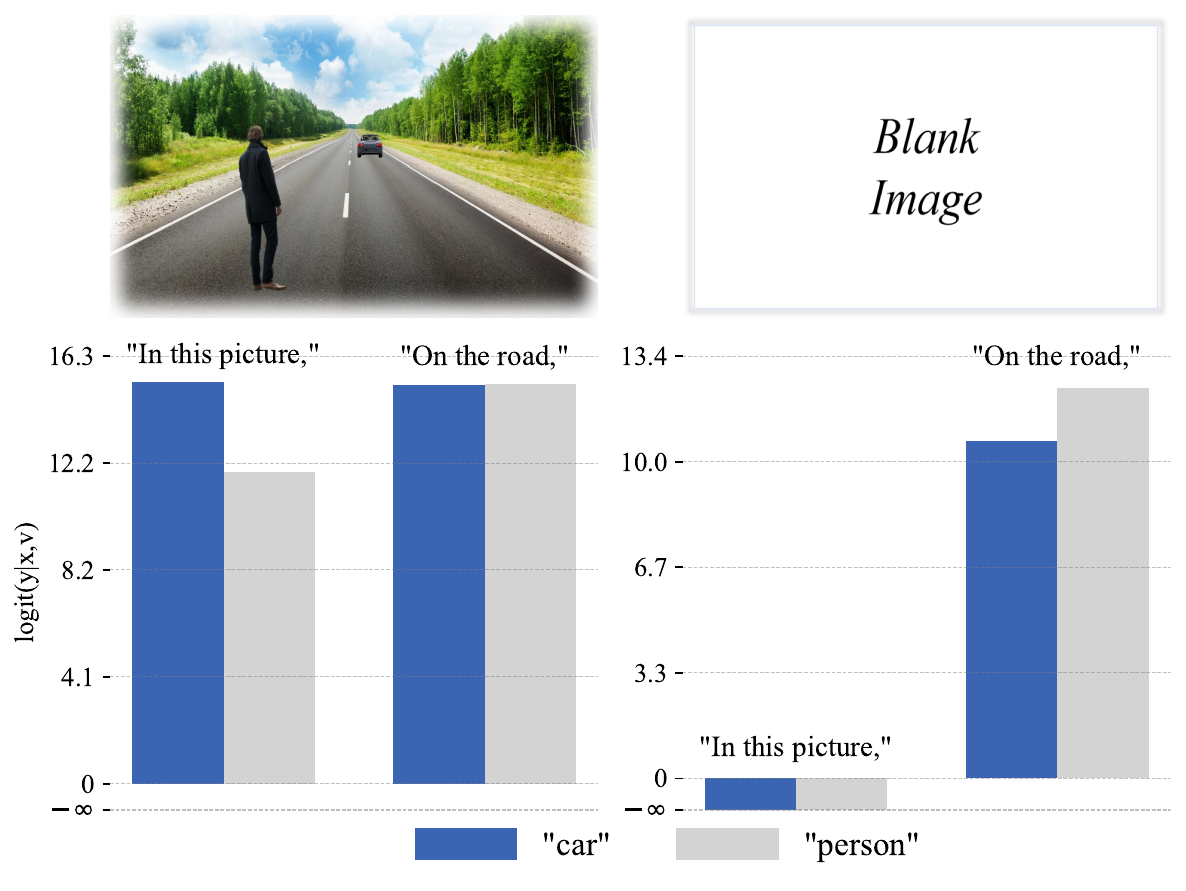}
  \caption{The figure above demonstrates the impact of different prompt inputs on the model's output. It explains the phenomenon of "Decoding Inertia" by introducing a blank image and two distinct prompt inputs.}
  \label{fig:figure3}
\end{figure}
The precision of the visual input is crucial to determine whether an LVLM generates hallucinations. However, the visual encoder exhibits lower sensitivity to entities that occupy a small area in the image \cite{liu2021survey}, often misrepresenting or failing to capture these small entities, leading to the generation of hallucinations. For LVLM, this phenomenon is also documented in \cite{Zhang2024ExploringPL}, where the uniform image resolution input setting of the visual encoder leads to the models lacking the same sensitivity to scale and detail as human vision. In the following, we will delve into the LVLM's sensitivity to entities with different area proportions in the image, and we will provide experimental evidence to support the validity of this hypothesis.

\noindent\textbf{Insensitivity to Local Details.} To demonstrate the model's sensitivity to local details, we designed an experiment in which three images of a "\textit{car}" were created at varying distances, meaning the "\textit{car}" progressively becomes smaller. A "\textit{person}" of constant size was used as a comparison. Figure \ref{fig:figure2} presents the three images, and we extracted the logits corresponding to the same query from the LVLM's feedback. The volume of the "\textit{person}" entity remains unchanged across all three images, and the corresponding logits show minimal variation. However, the logits for the "\textit{car}" entity almost linearly change in proportion to its size. This observation indicates a correlation between the LVLM's visual encoder sensitivity and the size of the entities in the image: \(logit_{i} \propto \frac{A_{i}}{A_{all}}\), (\(A \) represents the area of the region.) As shown in Equation \ref{eq:3}, visual information biases influence the model's reasoning process, which in turn exacerbates hallucinations caused by language priors.

\noindent\textbf{"Decoding Inertia": LVLM's Over-reliance on Language Priors.}  Interestingly, during the above experiment, we found that replacing the opening phrase "\textit{In this picture,}" with "\textit{On the road,}" in the prompt led to a significant shift in the LVLM's output distribution. As shown in Figure \ref{fig:figure3}, when we altered the prompt and tested the image with the smallest "\textit{car}" entity, the logit for the "\textit{car}" output significantly increased. To further investigate this phenomenon, we conducted an experiment with a blank image, using both prompt settings to query the LVLM. The logit results confirmed our hypothesis that the LVLM over-relies on language priors. When the word "\textit{road}" appeared, the probability of the "\textit{car}" entity also increased substantially. We refer to this phenomenon as "Decoding Inertia", which may be due to the frequent co-occurrence of specific objects in training data \cite{zhou2024analyzingmitigatingobjecthallucination,galleguillos2008object}.
\section{Method}
\subsection{Hallucination Disentangled Decoding}
\begin{figure*}[t]
    \centering
    \includegraphics[width=\textwidth, height=10cm]{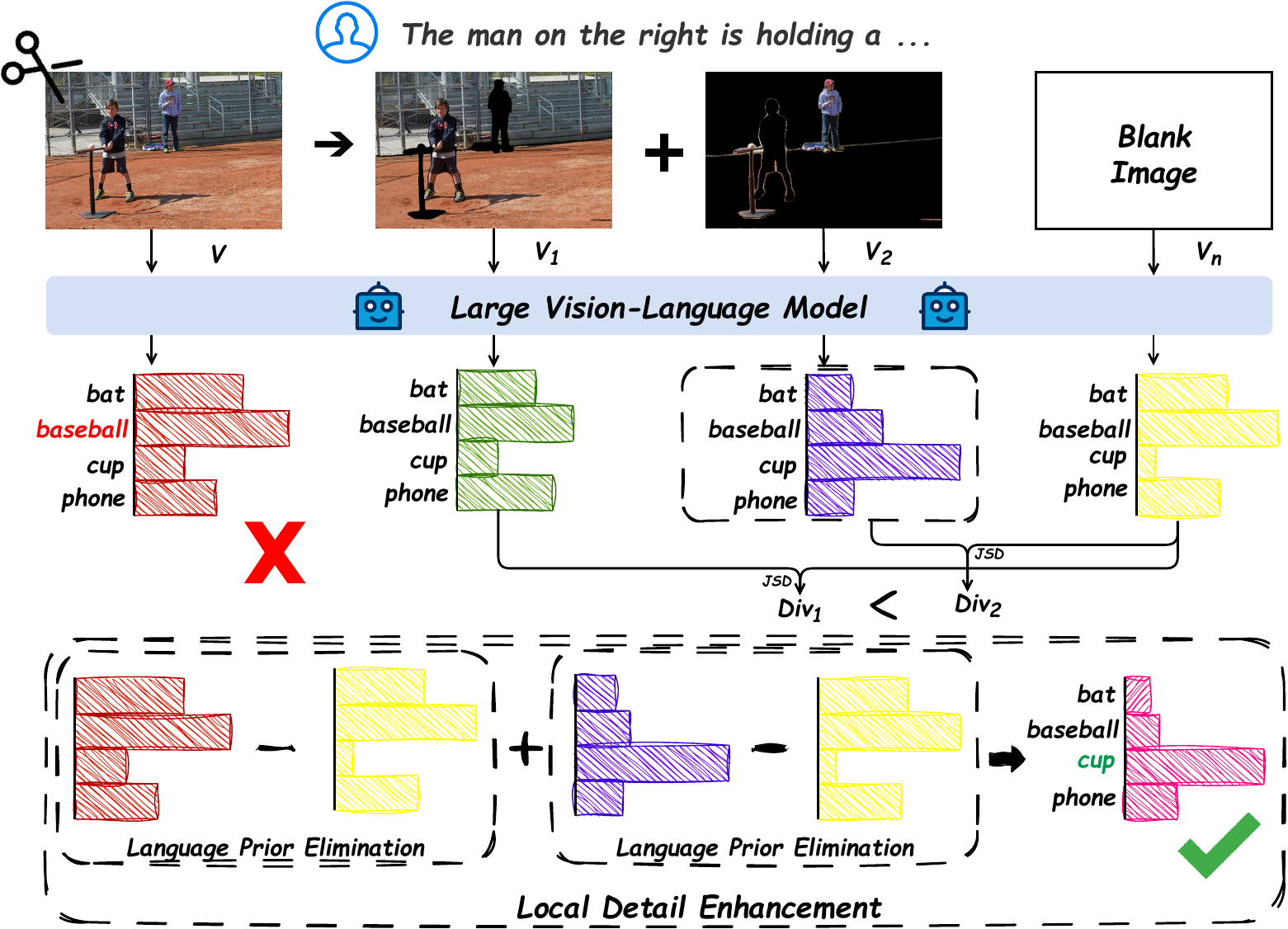}  
    \caption{This figure provides an overview of Hallucination Disentangled Decoding. First, the original image is segmented into two images. The blank image, along with the three aforementioned images, is then input into the LVLM, and their respective output distributions are obtained. By calculating the Jensen-Shannon (JS) divergence between the distributions of the two images and the blank image, the image with the higher JS divergence is selected. The original image and the chosen image are then subtracted from the blank image's distribution and summed to obtain the new output distribution.}
    \label{fig:overview}
\end{figure*}
From the experiments in the previous section, we found that both the vision module and the language module had shortcomings that could not be ignored, and the previous work was optimized for only one module. To address this issue, we propose Hallucination Disentangled Decoding (HDD), which aims to decouple and mitigate hallucinations in both the visual and language modules of the LVLM.
\subsubsection{Visual Detail Enhancement}
For hallucinations in the visual module, in order to amplify the proportion of local details in the image information, we use semantic segmentation model to segment the original image into two complementary images:
\begin{equation}
\label{eq:6}
\small
    v_{1}=\sum_{i=1}^N(Mask_i\odot \mathcal{V}), \textbf{ }
    v_{2}=\mathcal{V}-v_{1} 
\end{equation}
\(\mathcal{V}\) represent the original image, and \(v_{1}\) and \(v_{2}\) represent the two complementary segmented images. This approach effectively increases the proportion of the queried entity in the original image. The image segmentation strategy is selecting the largest \(N\) masks as the first image (if there is no special statement: \(N\!=\!0.05\!*\!n\)), and then masking the entities in the first image from the original, with the remaining entities forming the second image. Additionally, a completely blank image \(v_{n}\) is introduced. These four images: \(v\), \(v_{1}\), \(v_{2}\), and \(v_{n}\) are then used as inputs to the LVLM, generating four different output distributions. Our goal is to \textbf{automatically} select the most effective segmented image to answer the query. To achieve this, we use the Jensen-Shannon divergence \cite{fuglede2004jensen} to measure the differences between \(v_{1}\), \(v_{2}\) and \(v_{n}\).
\begin{equation}
\label{eq:7}
\small
\begin{split}
\mathcal{D}iv_{i} = \mathrm{JSD}\Big[p(y_t | v_{i}, x, y_{<t}) || p(y_t | v_{n}, x, y_{<t}) \Big]
\end{split}
\end{equation}
where \(i \! \in \! (1,2)\). A larger \(\mathcal{D}iv\) indicates a greater difference between the segmented image and the blank image, meaning that this segmented image contains more effective information. We use the segmented image with the larger \(\mathcal{D}iv\) to enhance the original image by adding the logit distribution of the selected segmented image to the logit distribution of the original image.

\noindent\textbf{Adaptive Weight Soft Adjustment} We also design an automatic weight adjustment mechanism to adaptively control the extent of local detail enhancement. From a mathematical perspective, we have:
\begin{equation}
\label{eq:8}
    \small
    \delta=\Big|\mathcal{D}iv_{1} - \mathcal{D}iv_{2}\Big|
\end{equation}
\(\delta\) allows the segmented image containing more effective information to provide stronger guidance to the original image, while conversely, the segmented image with less effective information exerts a weaker influence on the original image. Combining this with the previously discussed local detail enhancement, we have enhanced logit:
\begin{equation}
\small
    \begin{aligned}
    \mathrm{logit}_{enh}(\mathcal{V}) = (1 \! - \! \delta) \cdot \mathrm{logit}(\mathcal{V}) + \delta \cdot \mathrm{logit}(v_{i^*}&), \\
    \text{where } i^* = { \operatorname {arg\,max} } \  \mathcal{D}iv_{i}, \ i \in \{1,2\} &
    \end{aligned}
\end{equation}

\subsubsection{Language Prior Elimination}
\begin{table*}[t!]
\small
\centering
\begin{threeparttable}
\label{performance_comparison}
\setlength{\tabcolsep}{10pt}  
\begin{tabular}{c@{\hskip 4pt}c@{\hskip 4pt}ccccccc}
	\toprule
	\multirow{2}{*}{\textbf{Dataset}}& 
        \multirow{2}{*}{\textbf{Spilt}}& 
        \multirow{2}{*}{\textbf{Method}}& 
        \multicolumn{2}{c}{\textbf{LLaVA-1.5}}& 
        \multicolumn{2}{c}{\textbf{InstructBLIP}}& 
        \multicolumn{2}{c}{\textbf{LLaVA-NeXT}} \\
	\cmidrule(l){4-5} \cmidrule(l){6-7} \cmidrule(l){8-9}
&&&$\text{Acc.}\uparrow$&$\text{F1}\uparrow$&$\text{Acc.}\uparrow$&$\text{F1}\uparrow$&$\text{Acc.}\uparrow$&$\text{F1}\uparrow$ \cr
	\midrule
	\multirow{12}{*}{\rotatebox{90}{MSCOCO}}& \multirow{4}{*}{Random}& Greedy&88.7&87.3&87.0&86.3&89.3&88.8\cr
	&&+VCD&88.7&88.3&86.5&85.9&87.8&87.0\cr
        &&+SID&89.5&89.6&87.2&87.0&90.0&90.0\cr
	&&\textbf{+HDD}&\textbf{90.1}&\textbf{90.0}&\textbf{89.2}&\textbf{88.7}&\textbf{91.5}&\textbf{91.3}\cr
    \cmidrule(l){2-9}
    &\multirow{4}{*}{Popular}&Greedy&82.7&83.3&84.7&84.2&84.1&81.7\cr
	&&+VCD&87.3&87.1&84.3&83.8&87.7&87.6\cr
        &&+SID&87.1&87.2&85.1&85.4&88.2&87.5\cr
	&&\textbf{+HDD}&\textbf{88.0}&\textbf{88.0}&\textbf{87.0}&\textbf{86.8}&\textbf{89.2}&\textbf{89.0}\cr
    \cmidrule(l){2-9}
    &\multirow{4}{*}{Adversarial}&Greedy&81.1&80.7&80.0&80.1&83.2&81.0\cr
	&&+VCD&81.4&82.2&80.1&80.5&82.9&81.5\cr
        &&+SID&82.7&81.9&81.4&80.9&84.0&83.0\cr
	&&\textbf{+HDD}&\textbf{83.3}&\textbf{83.4}&\textbf{82.3}&\textbf{82.5}&\textbf{84.3}&\textbf{84.5}\cr
    
        \midrule
        
	\multirow{15}{*}{\rotatebox{90}{A-OKVQA}}& \multirow{5}{*}{\raisebox{-1em}{Random}} &Beam&86.3&87.4&88.3&88.4&86.6&85.3\cr
        &&+VCD&87.2&88.2&87.6&87.7&87.4&88.0\cr
	&&+OPERA&85.7&86.9&\textbf{89.0}&\textbf{89.4}&86.3&87.2\cr
        &&+HALC&88.2&87.4&88.4&87.7&89.0&88.2\cr
	&&\textbf{+HDD}&\textbf{89.5}&\textbf{89.4}&88.4&88.5&\textbf{89.5}&\textbf{89.6}\cr
    \cmidrule(l){2-9}
    &\multirow{5}{*}{Popular} &Beam&78.7&81.7&81.0&82.2&84.5&83.3\cr
        &&+VCD&79.5&82.6&80.8&81.9&84.8&86.7\cr
	&&+OPERA&79.9&82.5&79.5&81.8&84.3&85.0\cr
        &&+HALC&84.7&84.3&81.4&80.8&85.9&84.7\cr
	&&\textbf{+HDD}&\textbf{85.0}&\textbf{85.1}&\textbf{81.6}&\textbf{82.9}&\textbf{86.8}&\textbf{87.2}\cr
    \cmidrule(l){2-9}
    &\multirow{5}{*}{Adversarial} &Beam&68.1&74.9&74.7&77.9&78.6&78.3\cr
        &&+VCD&75.8&78.6&74.9&77.6&78.7&78.9\cr
	&&+OPERA&69.1&75.4&71.5&76.4&78.4&79.1\cr
        &&+HALC&77.5&79.6&74.9&77.2&78.7&79.2\cr
	&&\textbf{+HDD}&\textbf{78.0}&\textbf{79.9}&\textbf{75.7}&\textbf{79.0}&\textbf{79.1}&\textbf{79.5}\cr
    
        \midrule
        
        \multirow{12}{*}{\rotatebox{90}{GQA}}& \multirow{4}{*}{Random} &Sampling&83.5&82.7&79.5&80.4&83.1&81.2\cr
        &&+VCD&86.1&87.1&82.8&83.2&85.7&85.9\cr
	&&+RITUAL&86.8&87.2&83.5&84.3&88.7&88.9\cr
	&&\textbf{+HDD}&\textbf{87.2}&\textbf{87.6}&\textbf{84.1}&\textbf{84.9}&\textbf{89.3}&\textbf{89.4}\cr
    \cmidrule(l){2-9}
    &\multirow{4}{*}{Popular} &Sampling&78.1&78.3&73.5&76.1&78.2&77.1\cr
        &&+VCD&80.1&81.6&76.8&78.6&80.3&81.1\cr
	&&+RITUAL&79.5&79.8&76.7&78.7&84.4&86.2\cr
	&&\textbf{+HDD}&\textbf{80.5}&\textbf{81.3}&\textbf{77.0}&\textbf{80.1}&\textbf{86.8}&\textbf{87.3}\cr
    \cmidrule(l){2-9}
    &\multirow{4}{*}{Adversarial} &Sampling&74.8&75.8&70.4&74.1&75.4&74.7\cr
        &&+VCD&75.2&77.7&73.1&76.1&75.8&76.5\cr
	&&+RITUAL&75.3&76.1&72.8&75.5&77.2&78.3\cr
	&&\textbf{+HDD}&\textbf{76.4}&\textbf{78.1}&\textbf{73.5}&\textbf{76.0}&\textbf{78.4}&\textbf{80.5}\cr
	\bottomrule
\end{tabular}
\caption{The results of the POPE evaluation. We tested three different decoding strategies across various baselines and HDD. The best results are highlighted in bold.}
\label{tab:pope}
\end{threeparttable}
\end{table*}
In the previous section, we enhanced the capability of the LVLM's visual module. However, as demonstrated in the experiment shown in Figure \ref{fig:figure3}, both the original image and the segmented image in the LVLM can exhibit an over-reliance on language priors. To solve this problem, we introduce a blank image, serving as an image placeholder. This operation allows the LVLM to degrade into a simple language-only model, where the query input can elicit hallucinations based solely on language priors (since there is no visual input). Based on the analysis above, we perform contrastive decoding \cite{Wang2024MitigatingHI} by subtracting the output distribution of the blank image \(v_{n}\) from the output distribution of the image input, and add a weighting factor \(\alpha\). Mathematically, this can be formulated as:
\begin{equation}
\label{eq:10}
\small
    \mathrm{logit}^{\ast} (v_{in})=(1 \!+\! \alpha) \cdot \mathrm{logit}_{enh}(v_{in}) 
    \!-\!\alpha\\ \cdot \mathrm{logit}_{enh}(v_{n})
\end{equation}
Here, \(v_{in}\) includes the original image \(\mathcal{V}\) as well as the segmented images \(v_{1}\) and \(v_{2}\). Combining the local detail enhancement from the previous section, we finally have logits:
\begin{equation}
\label{eq:11}
\small
    \begin{aligned}  
        \mathbf{logit}_{hdd} = (1 - \delta) \cdot \mathrm{logit}^{\ast}(\mathcal{V}) + \delta \cdot \mathrm{logit}^{\ast}(v_{i^*}&),  \\
        \text{where } i^* = { \operatorname {arg\,max} } \,  \mathcal{D}iv_{i}, \ i \in \{1,2\}&
    \end{aligned}
\end{equation}
Our final logits as shown in Figure \ref{fig:overview} incorporate both the enhancement of the visual module and the elimination of language priors, effectively disentangling and mitigating hallucinations in both modules. This approach avoids compromising the performance of any module when reducing hallucinations, as well as other potential adverse effects.
\section{Experiments}
In this section, we evaluate the effectiveness of Hallucination Disentangled Decoding using multiple LVLMs and various benchmark tests.
\subsection{Experimental Settings}
\begin{table*}[t]
\small
\centering
\begin{threeparttable}
\label{performance_comparison}
\resizebox{1\textwidth}{!}{
\begin{tabular}{ccccccccccc}
	\toprule
	\multirow{2}{*}{\textbf{Method}}& \multicolumn{3}{c}{\textbf{LLaVA-1.5}}& \multicolumn{3}{c}{\textbf{InstructBLIP}}& \multicolumn{3}{c}{\textbf{LLaVA-NeXT}}&\cr
	\cmidrule(l){2-4} \cmidrule(l){5-7} \cmidrule(l){8-10}
	&$\text{CHAIR}_I\downarrow$&$\text{CHAIR}_S\downarrow$&$\text{Length}$&$\text{CHAIR}_I\downarrow$&$\text{CHAIR}_S\downarrow$&$\text{Length}$&$\text{CHAIR}_I\downarrow$&$\text{CHAIR}_S\downarrow$&$\text{Length}$\cr
	\midrule
        Multinomial Sampling&7.9&26.4&53.4&10.4&32.8&53.9&7.5&18.0&60.1\cr
        Greedy Decoding&6.0&21.0&54.7&7.3&26.2&55.4&6.0&14.8&60.7\cr
        Beam Search&5.9&20.8&54.5&7.1&25.8&55.3&5.7&14.4&60.4\cr
        \midrule
        VCD&6.4&22.4&54.3&\textbf{6.3}&24.0&54.4&6.1&15.1&60.2\cr
	OPERA&6.1&21.4&54.1&7.3&22.4&52.7&5.7&14.7&59.8\cr
        HALC&6.2&22.0&54.5&6.8&23.2&54.7&5.8&14.9&60.3\cr
        \textbf{HDD}&\textbf{5.6}&\textbf{18.4}&52.7&\textbf{6.3}&\textbf{21.2}&55.5&\textbf{5.5}&\textbf{13.6}&60.2\cr
	\bottomrule
\end{tabular}
}
\caption{The results of the CHAIR evaluation are presented. We tested various baselines and HDD on two different models, with all tests conducted under the condition of \textit{max new tokens = 64}.}
\label{tab:chair}
\end{threeparttable}
\end{table*}
\noindent\textbf{Evaluation Metrics}

\textbf{POPE} Polling-based Object Detection Evaluation \cite{Li2023EvaluatingOH} is a recently introduced method aimed at assessing hallucination issues in LVLMs. POPE focuses on evaluating object hallucinations by converting the hallucination assessment metric into a binary classification task using a question like "\textit{Is there a <object> in the image?}" to determine whether the model can correctly identify an object corresponding to a given image. The POPE test consists of three parts: The "random" setting randomly selects objects from the entire dataset for querying. The "popular" setting queries the most frequently occurring objects in the evaluation dataset. The "adversarial" setting queries objects that are highly relevant to the entities appearing in the image.

\begin{figure}[t]
  \includegraphics[width=\columnwidth]{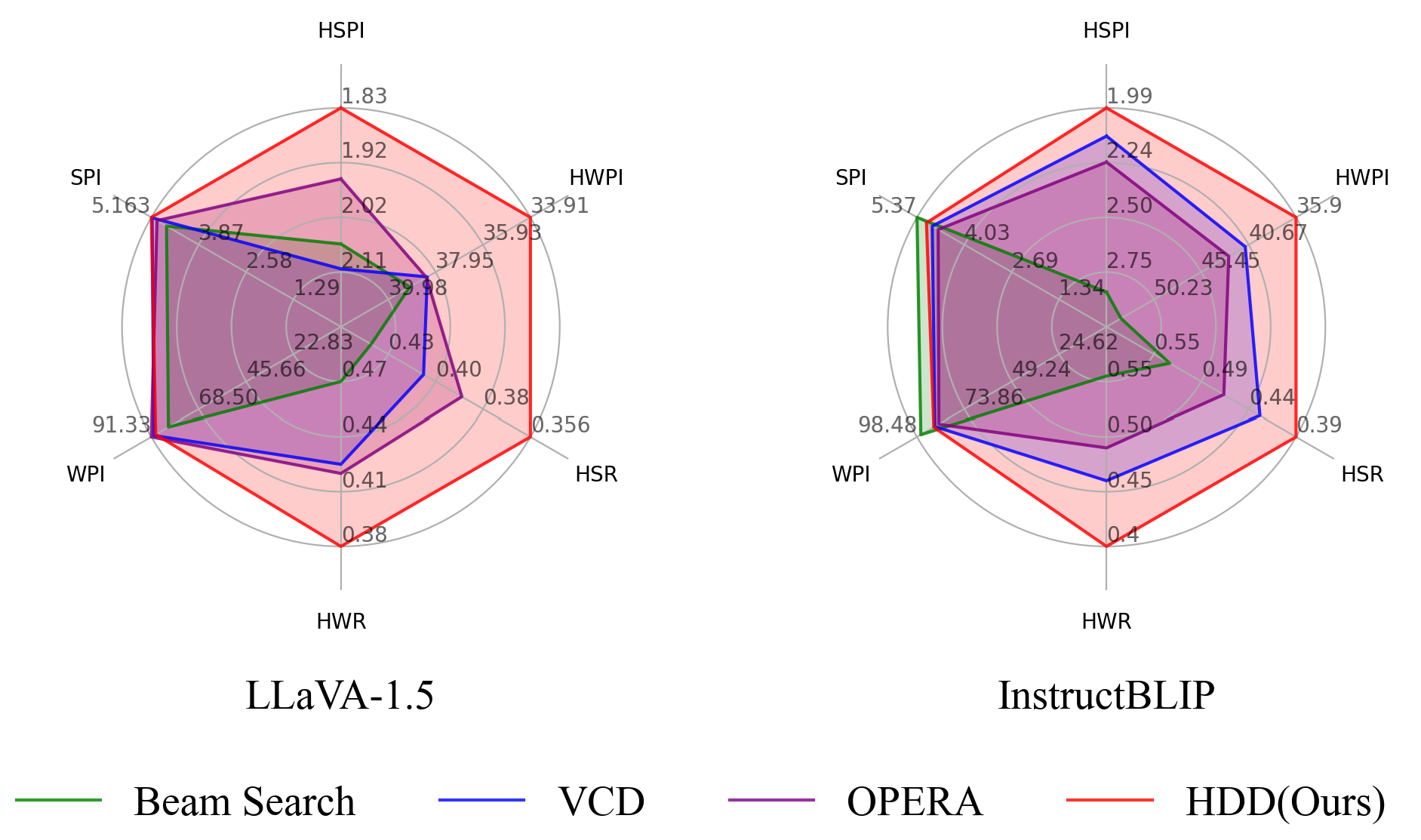}
  \caption{GPT-4 assisted hallucination evaluation results on VG dataset. We analyse the results across six dimensions, where HWR, HSR, HSPI, and HWPI are better when lower, while WPI and SPI are better when higher. Overall, a larger proportion on the radar chart indicates better performance.}
  \label{fig:rader}
\end{figure}
\textbf{CHAIR} Caption Hallucination Assessment with Image Relevance \cite{Rohrbach2018ObjectHI} is used to evaluate the accuracy of model outputs. This framework includes two evaluation dimensions: the sentence dimension (CHAIR\textsubscript{\textit{S}}) and the entity dimension (CHAIR\textsubscript{\textit{I}}). CHAIR\textsubscript{\textit{S}} represents the sentence-level evaluation, measuring the proportion of hallucinated sentences relative to all generated sentences, while CHAIR\textsubscript{\textit{I}} measures the object-instance level hallucination, indicating the proportion of hallucinated objects relative to all generated objects.For different settings, we used the same prompt: "\textit{Please describe this image in detail.}"

\textbf{GPT-4 Assisted Benchmark} POPE and CHAIR are both effective objective metrics, but they primarily focus on the object-existence-level hallucination and are unable to effectively capture hallucinated information related to entity relationships and spatial positions. Therefore, we introduce the GPT-4 Assisted Benchmark \cite{Zhao2023BeyondHE}, which leverages LVLM-generated detailed descriptions of the VG dataset to measure hallucinations in a more fine-grained manner. The benchmark using GPT-4 \cite{achiam2023gpt} and includes six evaluation dimensions: the number of sentences per image (SPI), the number of words per image (WPI), the number of hallucinated sentences per image (HSPI), the number of hallucinated words per image (HWPI), the ratio of hallucinated sentences (HSR) and the ratio of hallucinated wors (HWR). More details are in Appendix \ref{sec:appendixA}.

\noindent\textbf{Models and Datasets}

Based on prior work, we selected LLaVA-v1.5-7b \cite{Liu2023ImprovedBW}, InstructBLIP-7b \cite{Dai2023InstructBLIPTG} and LLaVA-NeXT \cite{li2024llavanext-strong}. Our evaluation metrics involve four different datasets: MS-COCO \cite{Lin2014MicrosoftCC}, A-OKVQA \cite{Schwenk2022AOKVQAAB}, GQA \cite{Hudson2019GQAAN} and Visual Genome (VG) dataset \cite{Krishna2016VisualGC}. 

\noindent\textbf{Baselines}

We compare HDD with several advanced decoding methods, including: VCD \cite{Wang2024MitigatingHI}, OPERA \cite{Huang2023OPERAAH}, HALC \cite{chen2024halcobjecthallucinationreduction}, SID \cite{huo2025selfintrospectivedecodingalleviatinghallucinations} and RITUAL \cite{Woo2024RITUALRI}. For all baselines, we performed a fair comparison using the hyperparameters recommended by each work.
\subsection{Experimental Results}
\begin{figure*}[t]
    \centering
    \includegraphics[width=\textwidth]{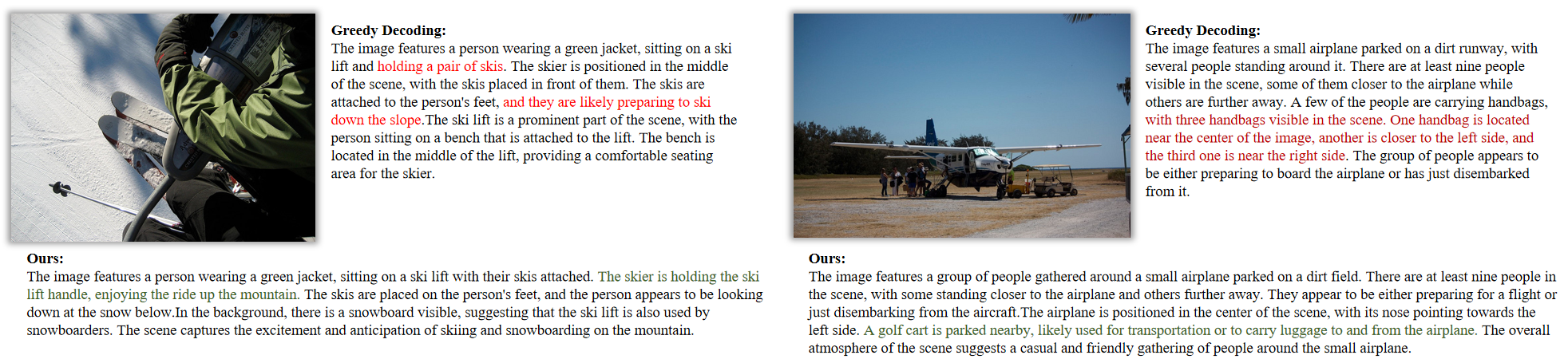}  
    \caption{The decoding differences on the MSCOCO dataset are presented, where the \textcolor{red}{Red} text highlights hallucinated regions, and the \textcolor{seagreen}{Green} text represents additional detailed descriptions not present in the original decoding. Our method not only reduces hallucinated information but also provides more detailed descriptions.}
    \label{fig:case}
\end{figure*}
\noindent \textbf{POPE Results} Our method is fully adaptable to various decoding strategies. The results of all POPE evaluations are presented in Table \ref{tab:pope}. It is evident that HDD consistently outperforms all baseline methods across almost all LVLMs, datasets, and decoding strategies (up to +9.9 Acc., +10.2 F1). In some cases, VCD and OPERA even show lower recognition accuracy than vanilla models, which is consistent with the results in \cite{huo2025selfintrospectivedecodingalleviatinghallucinations}. This suggests that our method not only eliminates hallucinations caused by language priors but also enhances the object recognition capabilities of LVLMs.

\noindent \textbf{CHAIR Results} For the CHAIR evaluation, we randomly selected 500 images from the MSCOCO dataset as the query set. As shown in the results from Table \ref{tab:chair}, our method achieves more accurate and detailed expression in real image recognition tasks, with both CHAIR\textsubscript{\textit{I}} and CHAIR\textsubscript{\textit{S}} outperforming all baseline results. Notably, our method significantly reduces CHAIR\textsubscript{\textit{I}} and CHAIR\textsubscript{\textit{S}} (with a maximum improvement of 29.1\% and 30.3\% respectively), indicating that HDD can more accurately recognize entities in the image and make more precise judgments for ambiguous entities.

\noindent \textbf{GPT-4 Benchmark Results} From the Figure \ref{fig:rader}, we can observe that HDD significantly outperforms all baselines on the hallucination metrics. It surpasses VCD by 12.9\% in the ratio of hallucinated sentences (HSR) and OPERA by 18.3\% in the ratio of hallucinated words (HWR). This indicates that our approach breaks through the bottlenecks of previous methods that only address hallucinations in a single module. Furthermore, as observed WPI and SPI, HDD eliminates hallucinations without sacrificing the detail of image descriptions.
\subsection{Further Discussions}
\begin{table}[t]
\centering
\begin{threeparttable}
\resizebox{1\columnwidth}{!}{
  \begin{tabular}{ccccccc}
    \toprule
    \multirow{2}{*}{\makecell{\textbf{Segmentation} \\ \textbf{Tool}}}& \multicolumn{2}{c}{\textbf{Random}}& \multicolumn{2}{c}{\textbf{Popular}}& \multicolumn{2}{c}{\textbf{Adversarial}} \\
	\cmidrule(l){2-3} \cmidrule(l){4-5} \cmidrule(l){6-7}
    &Acc.$\uparrow$ & F1$\uparrow$ &Acc.$\uparrow$ & F1$\uparrow$ & Acc.$\uparrow$ & F1$\uparrow$ \\
    \midrule
   \textbf{SAM}&\textbf{90.1}&\textbf{90.0}&\textbf{88.0}&\textbf{88.0}&83.3&\textbf{83.4}\\
   \textbf{Mask2Former}&90.0&89.9&87.9&87.5&\textbf{83.5}&83.2 \\
   \textbf{Mask R-CNN}&89.8&89.9&87.5&87.4&83.0&83.1\\
    \bottomrule
  \end{tabular}
  }
  \caption{Ablation study of segmentation tools}
  \label{tab:segmentatoin}
  \end{threeparttable}
\end{table} 
\noindent\textbf{Impact of Segmentation Tools} To demonstrate that our method is not dependent on a specific segmentation tool and is robust across different semantic segmentation models, we conduct an ablation study by replacing the primary segmentation backbone SAM with Mask2Former and Mask R-CNN. We perform POPE evaluation on the MSCOCO dataset using LLaVA-1.5 as shown in Table \ref{tab:segmentatoin}. Experimental results show that the choice of segmentation model has minimal impact on the final performance, with the largest accuracy difference being only 0.5\%. This slight variation may stem from differences in segmentation granularity across models.

\begin{figure}[t]
  \includegraphics[width=\columnwidth]{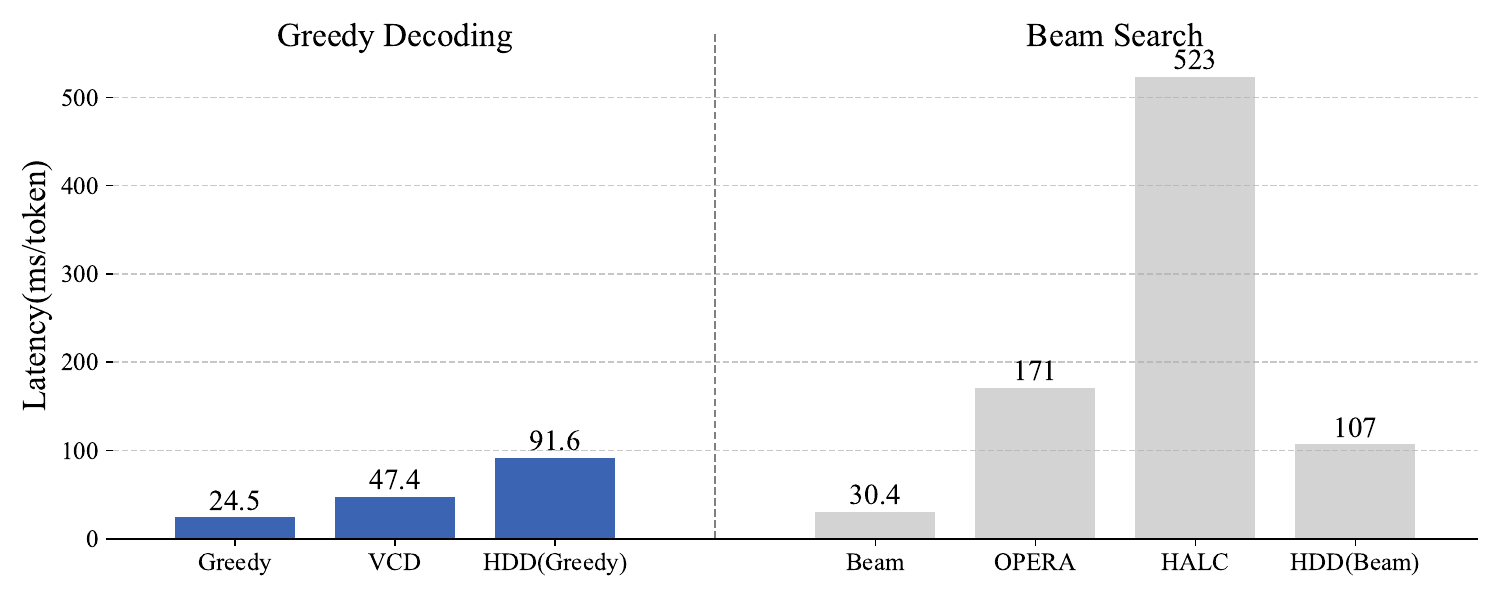}
  \caption{The decoding delay of different decoding methods, in ms/token. Higher values mean lower throughput.}
  \label{fig:latency}
\end{figure}

\noindent\textbf{Computational Efficiency Comparison} We measure the computational efficiency of different methods by the time each model requires to generate a single token. All experiments were conducted on a single NVIDIA RTX 4090 GPU. We sample inference times of LLAVA-1.5 on MSCOCO from the Chair evaluation and compute the corresponding output latencies. As shown in Figure \ref{fig:latency}, HDD operates within an acceptable efficiency range while demonstrating its effectiveness across various evaluations. HDD achieves latency comparable to VCD and base decoding methods, and significantly outperforms HALC and OPERA. These results indicate that HDD is a practical and efficient approach suitable for real-world scenarios.

\noindent\textbf{Case Study on MSCOCO} Figure \ref{fig:case} presents two samples of LLaVA-1.5 on MSCOCO. The red regions represent hallucinated information, while the green regions represent detailed information. Our method not only eliminates hallucinated information but also describes finer details in the image. HDD successfully identifies "\textit{ski lift handle}" and "\textit{golf cart}" and accurately analyzes the scene's intentions, such as "\textit{The skier is holding the ski lift handle, enjoying the ride up the mountain.}" and "\textit{A golf cart is parked nearby, likely used for transportation or to carry luggage to and from the airplane.}" These conclusions are drawn from observing the details of the scene.

\section{Conclusion}
In this paper, we proposes a disentangled solution called Hallucination Disentangled Decoding (HDD). Through a fine-grained analysis of the sources of hallucinations in both visual and language modules of LVLMs, we design two mechanisms to mitigate hallucinations in each module separately. (1) Using semantic segmentation model to segment images and enhance local details. (2) Using contrastive decoding with blank images to eliminate hallucinations caused by language priors. Our method outperforms existing methods in multiple benchmark tests and demonstrates superior robustness and generalizability.
\section{Limitation}
Although our approach achieves excellent results across various evaluations, there is still room for improvement. Our current research focuses on cross-modal alignment between images and text, without extending to other modalities such as video or 3D point clouds. While hallucinations remain prevalent in these modalities, we leave the exploration of applying our proposed method to mitigate hallucinations in broader multimodal settings as future work.
\clearpage

\bibliography{custom}

\clearpage
\appendix
\section{Implementation Details}
\label{sec:appendixA}
\textbf{POPE Setting}

For the POPE evaluation, 500 images are randomly selected from each of MS-COCO, AOKVQA and GQA. Each dataset is divided into three subsets based on different questioning strategies: Random, Popular, and Adversarial. In each subset, each image is paired with six distinct queries, resulting in a total of 3000 queries. The three datasets together contain a total of 27,000 queries. The hyperparameter $\alpha$ is set in the range of 0.1 to 0.6 for LLaVA-1.5 and InstructBLIP, while for LLaVA-NeXT, it is adjusted within the range of 1.0 to 1.6. The \(N\) in Equation \ref{eq:6} is set to \(0.05*n\) in Equation \ref{eq:5}. Beam Search is employed with a beam size of 2, and the temperature is set to 1 under multinomial decoding. OPERA, being designed based on beam search, does not produce results under greedy decoding and multinomial decoding settings.

\textbf{CHAIR Setting}

For the CHAIR evaluation, we randomly select 500 images from the MS-COCO dataset and generate image captions using our method as well as other decoding methods for evaluation. For fairness, VCD, OPERA, HALC and our HDD are all evaluated under the beam search setting with respective default settings, and their performance is compared with the original model using different decoding strategies, including greedy decoding, beam search, and multinomial sampling. The \(\alpha\) is set within the range of 0.1 to 0.6, the \(N\) in Equation \ref{eq:6} is set to \(0.05*n\) in Equation \ref{eq:5}, the \textit{max new tokens} is set to 64 across all methods and the temperature is set to 1 under multinomial decoding.

\textbf{GPT-4 Assisted Benchmark Setting}

We randomly selected 200 images from the VG dataset and used the prompt `\textit{Please describe the image in detail.}' to obtain the model's description of the images. The descriptions were then evaluated by GPT-4 for fine-grained assessment. Our GPT-4 Assisted Benchmark is based on HalluBench \cite{Zhao2023BeyondHE}. The detailed prompt settings will be presented in Table \ref{tab:gpt4-prompt}. For fairness, VCD, OPERA, and our HDD are all evaluated under the beam search setting. The \(\alpha\) is set within the range of 0.1 to 0.6, the \(N\) in Equation \ref{eq:6} is set to \(0.05*n\) in Equation \ref{eq:5}, the \textit{max new tokens} is set to 512 across all methods and the temperature is set to 1 under multinomial decoding.

\section{Adaptive Plausibility Constraints}
\label{sec:appendixB}
In Equation \ref{eq:3}, the segmented images \(v_{i}\) perform visual enhancement on the original image, but these segmented images are not always reasonable in every situation. This uncertainty amplifies the uncertainty in the model's output distribution, ultimately leading to implausible outputs. To address this issue, We implemented an adaptive consistency constraint following \cite{Li2022ContrastiveDO}, which depends on the confidence level associated with the output distribution relative to the original visual input. The mathematical form is as follows:
\begin{equation}
\label{eq:12}
\small
\begin{aligned}
    \mathcal{C}_{head}(y_{<t})=\{y_t\in \mathcal{C}:p(y_t|v,x,y_{<t}) \geq& \\ \beta \max_{w} p(w|v,x,y_{<t})&\}
\end{aligned}
\end{equation}
where \(\beta\) is a hyperparameter in the range of [0,1], we set \(\beta\) to 0.1 across all experiments.

\section{Ablation Studies}
\label{sec:appendixD}
\subsection{Distribution of \(\delta\)}
\begin{figure}[h]
  \includegraphics[width=\columnwidth]{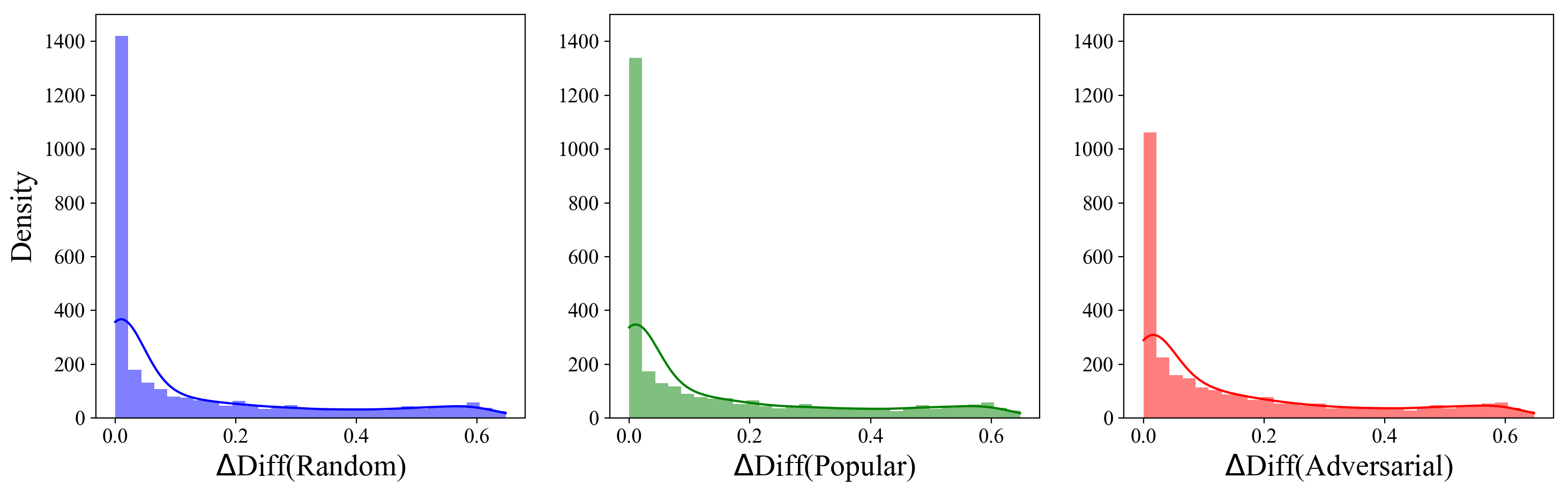}
  \caption{This figure illustrates the distribution of \(\delta\) for LLaVA-1.5 on MS-COCO. The three subplots represent the distribution of \(\delta\) across the Random, Popular, and Adversarial subsets.}
  \label{fig:diff}
\end{figure}
The values of \(\delta\) are mainly distributed within the range of 0 to 0.6, with nearly half of the \(\delta\) values concentrated near 0 (each subset contains 3000 queries). This occurs because, in the POPE evaluation, half of the queries have the answer 'No'. When such queries, along with two segmented images, are input into LVLMs, both segmented images should have a logit distribution similar to that of a blank image (since neither image contains the object mentioned in the query), resulting in a \(\delta\) close to 0. Compared to the Random subset, the Popular and Adversarial subsets show fewer values near 0, as the entities queried in the Popular and Adversarial subsets are more likely to induce hallucinations in the LVLM. This Figure \ref{fig:diff} essentially represents the distribution of positive and negative samples in the dataset.
\subsection{Effect of Parameter \(\alpha\)}
\begin{table}[h]
\centering
  \begin{tabular}{lccc}
    \toprule
    $\textbf{\(\alpha\) }$ & $\textbf{ACC}\uparrow$ & $\textbf{F1}\uparrow$ & $\textbf{Yes Ratio(\%)}$ \\
    \midrule
   0.2  & $88.1_{(+4.8)}$ & $87.4_{(+6.1)}$ & 44.6  \\
   0.4  & $88.9_{(+5.6)}$ & $88.6_{(+7.3)}$ & 47.1 \\
   0.6  & $89.3_{(+6.0)}$ & $89.2_{(+7.9)}$ & 49.2 \\
   0.8  & $89.1_{(+5.8)}$ & $89.2_{(+7.9)}$ & 50.4 \\
   1.0  & $88.9_{(+5.6)}$ & $89.0_{(+7.7)}$ & 51.7 \\
   \midrule
   vanilla  & 83.3 & 81.3 & 39.7 \\
    \bottomrule
  \end{tabular}
  \caption{Ablation study of \(\alpha\) on the POPE benchmark}
  \label{tab:alpha}
\end{table} 
Table \ref{tab:alpha} presents the POPE evaluation results with different alpha values in Equation \ref{eq:10} (showing results for LLaVA-1.5 on the MS-COCO random subset, using multinomial sampling with a temperature setting of 1). We selected a range from 0.2 to 1.0 with a step size of 0.2. The impact of different alpha settings on the results is minimal, with stable results across the range of alpha from 0.2 to 1.0, all significantly outperforming the baseline setting. This demonstrates that our method exhibits good consistency and stability across different alpha settings.
\subsection{CHAIR Recall Analysis}
\begin{figure}[h]
  \includegraphics[width=\columnwidth]{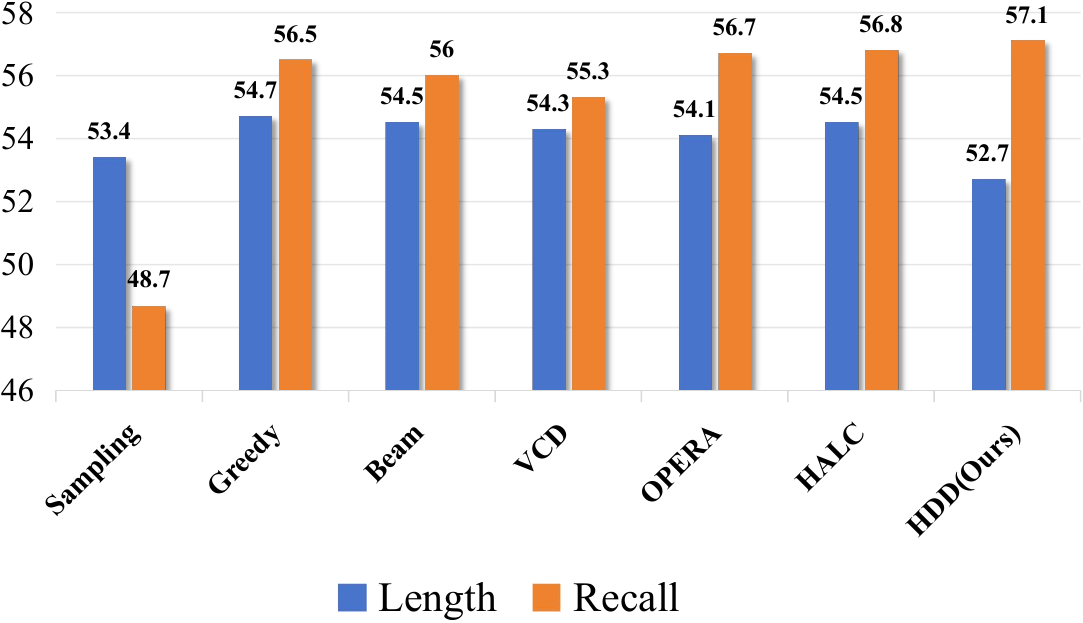}
  \caption{This chart illustrates the Recall and Length metrics of various decoding methods in the CHAIR evaluation for LLaVA-1.5.}
  \label{fig:recall}
\end{figure}
Building on CHAIR, we followed the design of \cite{yue-etal-2024-less} to compute the Recall values for each setting, in order to assess the semantic richness of the model's generated statements. As shown in the results in Figure \ref{fig:recall}, With a similar generation length, HDD achieved the highest recall value, surpassing the default decoding settings by 17.2\%. Combining the CHAIR evaluation results in Table \ref{tab:chair}, our HDD generates more effective information and fewer hallucinations per unit length of description.
\section{More Case Studies}
\label{sec:appendixE}
\begin{figure*}[t]
  \includegraphics[width=\textwidth]{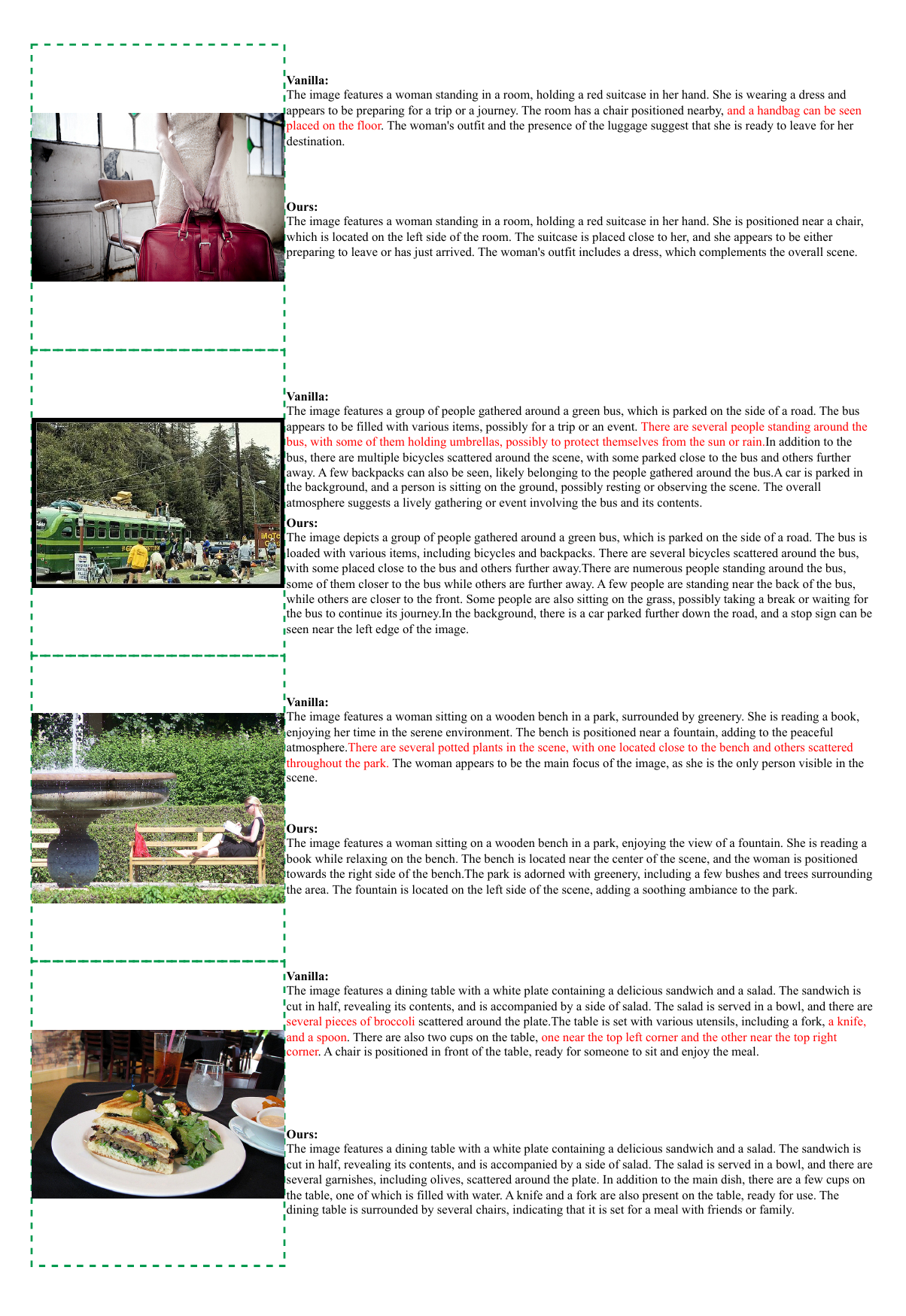}
  \caption{This figure presents more examples of LLaVA-1.5 on MS-COCO, with hallucinated information highlighted in \textcolor{red}{Red}.}
  \label{fig:more_case}
\end{figure*}

More case studies are shown in Figure \ref{fig:more_case}.

\begin{table*}
\begin{tabularx}{\textwidth}{
    >{\hsize=1.0\hsize\raggedright\arraybackslash}X 
    >{\hsize=2.0\hsize\centering\arraybackslash}X 
    >{\hsize=0.5\hsize\raggedleft\arraybackslash}X
}
\toprule
\textbf{GPT-4 Assisted Benchmark Prompt} \\
\midrule
Please help me judge if the comment of this image is hallucination or correct. \\
I will give you a list of region description of a image. The format is [x1, y1, x2, y2]: region description, where [x1, y1, x2, y2] is the bounding box of the region. Highly overlapping bounding boxes may refer to the same object. This is the ground truth information of the image. Besides, I give you some factual information about the content of the image (which is 100\% accurate). Your judgement should base on this information. However, this information only descibe the objects in the region of image, so it cannot descibe the subjective part of the image, e.g., atmosphere, style, emotion. In that case, you can return "Cannot judge". \\
Also, I will give you a list of comments of the image for you to judge if it is hallucination. Please give a judgement one by one along with the reason. \\
 \\
Your output should be: \\
Judgement: \\
1. hallucination or correct or cannot judge: <reason> \\
2. ... \\
 \\
Here are the region descriptions of the image: \\
\{\} \\
 \\
Factual Information: \\
\{\} \\

Here is the comment for you to judge (hallucination, correct, or cannot judge): \\
\{\} \\
\bottomrule
\caption{The prompt used for GPT-4 Assisted Benchmark}
\label{tab:gpt4-prompt}
\end{tabularx}
\end{table*}

\end{document}